# AN AUTOMATIC ALGORITHM FOR OBJECT RECOGNITION AND DETECTION BASED ON ASIFT KEYPOINTS

Reza Oji

Department of Computer Engineering and IT, Shiraz University
Shiraz, Iran
oji.reza@gmail.com

## ABSTRACT

*Object recognition is an important task in image processing and computer vision. This paper presents a perfect method for object recognition with full boundary detection by combining affine scale invariant feature transform (ASIFT) and a region merging algorithm. ASIFT is a fully affine invariant algorithm that means features are invariant to six affine parameters namely translation (2 parameters), zoom, rotation and two camera axis orientations. The features are very reliable and give us strong keypoints that can be used for matching between different images of an object. We trained an object in several images with different aspects for finding best keypoints of it. Then, a robust region merging algorithm is used to recognize and detect the object with full boundary in the other images based on ASIFT keypoints and a similarity measure for merging regions in the image. Experimental results show that the presented method is very efficient and powerful to recognize the object and detect it with high accuracy.*

## 1. INTRODUCTION

Extracting the points from an image that can give best define from an object in image namely keypoints is very important and valuable. These points have many applications in image processing like object detection, object and shape recognition, image registation and object tracking. By extracting the keypoints, we can use them for finding objects in the other images. Detect and recognize the object by using the keypoints and a segmentation algorithm is very accurate because if the keypoints are correctly identified, they achieve the best information from the image. ASIFT is a fully affine invariant method with respect to six parameters of affine transform [1,2], wherease the previous method SIFT was invariant with respect to four parameters namely translation, rotation and change in scale (zoom) [3]. ASIFT cover two other parameters namely longitude and latitude angle that are relevant to camera axis orientation. It means that ASIFT is more effective for our goal and can be more robust in the changes of images.

Many image segmentation and object recognition algorithms have been presented, each having its own specifications [4,5]. Some of these algorithm are interactive image segmentation based on region merging [6,7]. One of them is a powerful algorithm [8] for detecting object and its boundary but it is not an automatic algorithm and has a problem. In this algorithm the users must indicate some of locations and regions of the background and object to run the algorithm.

 29

In this paper, we combined ASIFT results and a region merging algorithm to recognize objects in images and detect them with full boundary. The presented algorithm does not have the stated problem in region merging algorithm. It means that, the algorithm does not need to indicate regions by user. We use the best keypoints of object that has been obtained from ASIFT results and apply them into the image. Therefore, the method will be an automatic algorithm and will not need marks by users and the achieved keypoints from ASIFT have been replaced with them. Lowe presented an object recognition algorithm [9] based on SIFT, but object recognition and detection algorithm with full boundary by using ASITF, has not yet been presented. Therefore, at this time we have an automatic algorithm for object recognition and full detection by using the obtained keypoints from ASIFT algorithm. These keypoints are very efficient and give us the best information of object to find it in the other images.

## 2. THE ASIFT ALGORITHM

ASIFT is an improved algorithm from SIFT. SIFT has been presented by Lowe (1994) where SIFT is invariant to four of the six parameters of affine transform. ASIFT simulates all image views obtained by the longitude and the latatude angle (varying the two camera axis orientation parameters) and then applies the other parameters by using the SIFT algorithm. There is a new notion called transition tilt that is designed to quantify the amount of tilt between two such images. In continuation, major steps of the algorithm are described.

### 2.1. The Affine Camera Model

Image distortions coming from viewpoint changes can be modeled by affine planar transforms. Any affine map A is defined:

$$A = \lambda \begin{bmatrix} \cos\psi & -\sin\psi \\ \sin\psi & \cos\psi \end{bmatrix} \begin{bmatrix} t & 0 \\ 0 & 1 \end{bmatrix} \begin{bmatrix} \cos\varphi & -\sin\varphi \\ \sin\varphi & \cos\varphi \end{bmatrix} \qquad (1)$$

Which we note $A = \lambda R_1(\psi)T_t R_2(\varphi)$, where $\lambda > 0$, $\lambda t$ is the determinant of A, $R_i$ are rotations, $\varphi \in [0,180)$, and $T_t$ is a tilt namely a diagonal matrix with first eigenvalue $t > 1$ and the second one equal to 1. Figure1, shows a camera motion interpretation of (1). $\theta$ and $\varphi = \arccos 1/t$ are respectively the camera optical axis longitude and latitude, $\lambda$ corresponds to the zoom and a third angle $\psi$ parameterizes the camera spin.

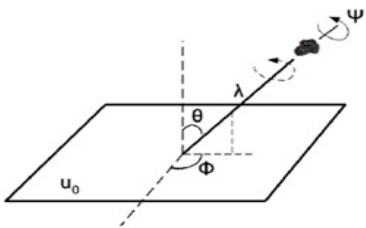

Figure 1. Camera motion interpretation

According to up preamble, each image should be transformed by simulating all possible affine distortions caused by the changes of camera optical axis orientations from a frontal position. These distortions depend upon two parameters, the longitude and the latitude. rotations and tilts are performed for a finite number of angles to decrease complexity and time computation. The sampling steps of these parameters ensuring that the simulated images keep close to any other possible view generated by other values of $\varphi$ and $\theta$.

After this process, the SIFT algorithm will be applied in all simulated images. The other steps are relevant to SIFT.

## 2.2. Scale Space Extremum Detection

The next step is extracting keypoints that are invariant to changes of scale. We need to search for stable features in all possible changes. In previous research it has been shown that for this task, the Gaussian function is the only possible scale-space kernel. A function, L(x,y,σ), is the scale-space of an image that is obtained from the convolution of an input image, I(x,y), with a variable scale-space Gaussian function, G(x,y,σ). For efficiently detect stable keypoint locations, Lowe proposed using the scale-space extremum in difference-of-Gaussian (DOG) function, D(x,y,σ). Extremum computed by the difference of two nearby scales separated by a constant factor K (2):

$$G\left(x,y,\sigma\right)=\frac{1}{2\pi\sigma^2}e^{-\left(x^2+y^2\right)/2\sigma^2} \quad , \quad \begin{aligned} D\left(x,y,\sigma\right)&=\left(G\left(x,y,k\sigma\right)-G\left(x,y,\sigma\right)\right)*I\left(x,y\right)\\ &=L\left(x,y,k\sigma\right)-L\left(x,y,\sigma\right) \end{aligned} \quad (2)$$

This process repeats in several octaves. In each octave, the initial image is repeatedly convolved with Gaussians to produce the set of scale space images. Adjacent Gaussian images are subtracted to produce the difference-of-Gaussian images. After each octave, the Gaussian images are down-sampled by a factor of 2, and the process repeated. Figure 2, shows a visual representation from this process.

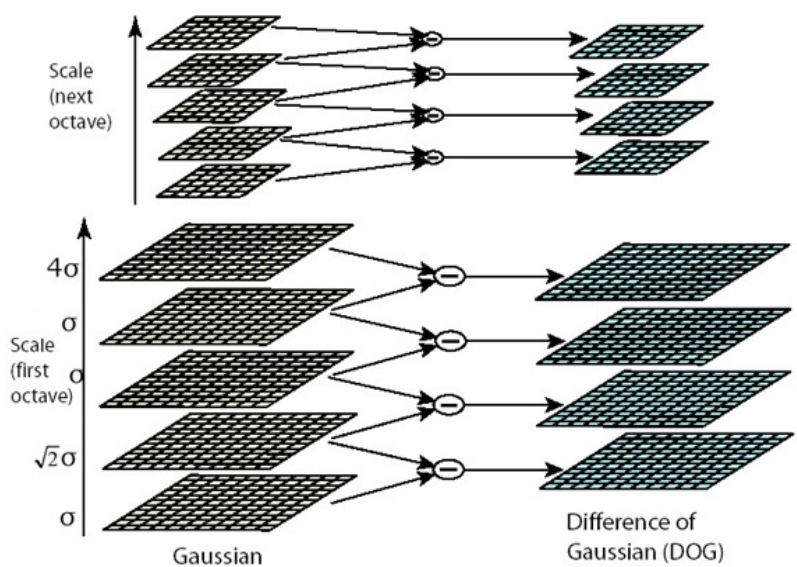

Figure 2. Visual representation of DOG in octaves

Each sample point (pixel) is compared with its neighbors according to their intensities for finding out whether is smaller or larger than neighbors. For more accuracy, each pixel will be checked with the eight closest neighbors in image location and nine neighbors in the scale above and below (Figure 3). If the point is an extremum against all 26 neighbors, is selected as candidate keypoint. The cost of this comparison is reasonably low because most sample point will be eliminated at first few checks.

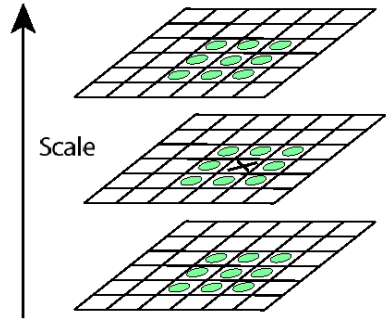

Figure 3. Local extremum detection of DOG

## 2.3. Accurate Keypoint Localization

For each candidate keypoint interpolation of nearby data is used to accurately determine its position. Then, many keypoints that are unstable and sensitive to noise, like the points with low contrast and the points on the edge, will be eliminated. (Figure 4)

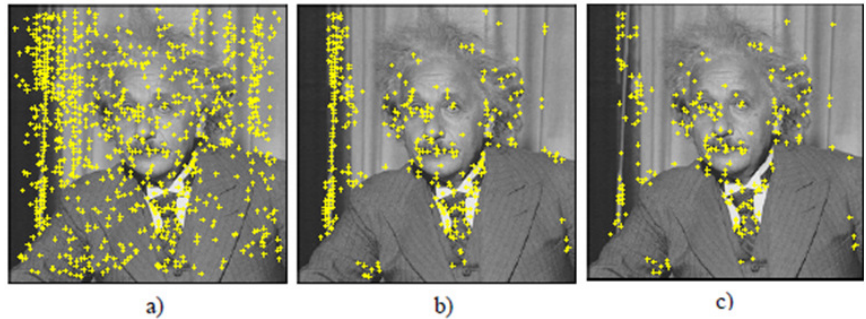

Figure 4. (a) Extremum of DOG across scales. (b) Remaining keypoints after removal of low contrast points. (c) Remaining keypoints after removal of edge responses.

## 2.4. Assigning an Orientation

The next step is assigning an orientation for each keypoint. By this step, the keypoint descriptor [10] can be represented relative to this orientation and therefore get invariance to image rotation. For each Gaussian smoothed image sample, points in regions around keypoint are selected and magnitude, m, and orientations, θ, of gradiant are calculated (3):

$$m(x, y) = \sqrt{(L(x+1, y) - L(x-1, y))^2 + (L(x, y+1) - L(x, y-1))^2}$$

$$\theta(x, y) = \tan^{-1}((L(x, y+1) - L(x, y-1))/(L(x+1, y) - L(x-1, y))) \quad (3)$$

Then, created weighted (magnitude + gaussian) histogram of local gradiant directions computed at selected scale. Histogram is formed by quantizing the orientations into 36 bins to covering 360 degree range of orinetations. The highest peak in the histogram is detected where peaks in the orientation histogram correspond to dominant directions of local gradiant. (Figure 5)

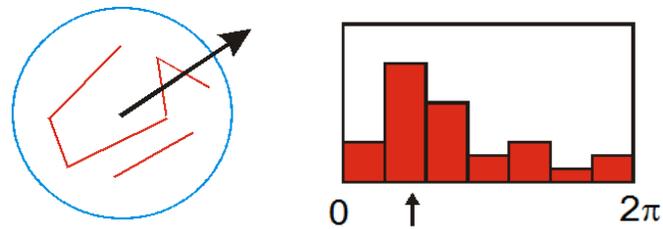

Figure 5. Visual representation of orientation histogram.
(the histogram is shown in 7 bins)

## 2.5. Keypoint Descriptor

The previous operations have assigned location, scale, and orientation to each keypoint and provided invariance to these parameters. Remaining goals are to define a keypoint descriptor [11] for the local image regions and reduce the effects of illumination changes.

The descriptor is based on 16×16 samples that the keypoint is in the center of. Samples are divided into 4×4 subregions in 8 directions around keypoint. Magnitude of each point is weighted and gives less weight to gradients far from keypoint (Figure 6). Therefore, feature vector dimensional is 128 (4×4×8).

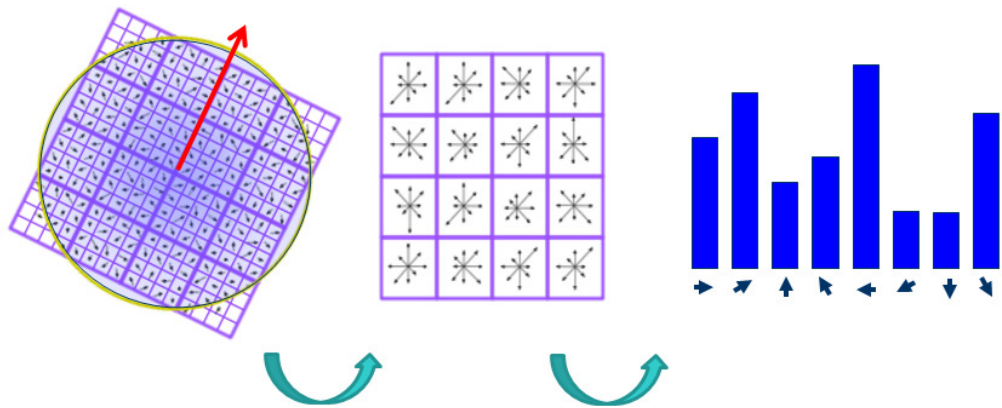

Figure 6. A keypoint descriptor

Finally, vector normalization is applied. The vector is normalized to unit length. A change in image contrast in which each pixel value is multiplied by a constant will multiply gradients by the same constant. Contrast change will be canceled by vector normalization and brightness change in which a constant is added to each image pixel will not affect the gradient values, as they are computed from pixel differences.

Now we can find keypoints from an image in the other images and match them together. One image is the training sample of what we are looking for and the other image is the world picture that might contain instances of the training sample. Both images have features associated with them across different octaves. Keypoints in all octaves in one image independently match with all keypoints in all octaves in other image. Features will be matched by using nearest neighbor algorithm. The nearest neighbor is defined as the keypoint with minimum Eculidean distance for the invariant descriptor vector as described upside. Also to solve the problem of features that have no correct match due to some reason like background noise or clutter, a threshold at 0.8 is chosen for ratio of closest nearest neighbor with second closest nearest neighbor that obtained experimentally . If the distance is more than 0.8, then the algorithm does not match keypoints together.

# 3. PROPOSED OBJECT RECOGNITION AND DETECTION METHOD

For each object, we train it by ASIFT algorithm in several images with different scales and viewpoints to find best keypoints to recognize desire object in the other images. Then, by using these keypoints and a region merging algorithm the object will be detected. A maximal similarity region mering segmentation algorithm was propoesd by Ning. This algorithm is based on a region merging method with using initial segmentation of mean shift [12] and user markers. User marks part of object and background to help the region merging process by maximal similarity. The non-marker background regions will be automatically merged while the non-marker object regions will be identified and avoided from being merged with background. It can extract the object from complex scenes but there is a weakness that the algorithm needs user marks, hence, the algorithm is not automatic. In our algorithm, we use this region merging but without user marks. We propose an efficient method for object recognition and detection by combining ASIFT keypoints (instead of user marks) with an automatic segmentation based on region merging which can detect object with full boundary.

## 3.1. Region Merging Based on Maximal Similarity

An initial segmentation is required to partition image into regions for region merging in further steps. We use the mean shift segmentation software namely the EDISON System [13] for initial segmentation because it can preserve the boundries well and has high speed. Any other low level segmentation like super-pixel and watershed can be used too. Please refer to [14,15] for more information about mean shift segmentation.

There is many small region after initial segmentation. These regions represent using color histogram that is an effective descriptor because different regions from the same object often have high similarity in color whereas there have variation in other aspects like size and shape. The color histogram computed by RGB color space. We uniformly quantize each color chanel into 8 levels. Therefore, the feature space is 8×8×8=512 and the histogram of each region is computed in 512 bins. The normalized histogram of a region X denote by $Hist_X$. Now we want to merge the regions by their color histogram that can extract the desired object.

When we applied ASIFT keypoints in the image, some regions in image that are relevant to object in image are compoesd of keypoints and the others are not. Here, an important issue is how to define the similarity between the regions inclusive keypoints with the regions without any point so that the similar regions can be merged. We use the Euclidean distance for this target because that is a well known goodness of fit statistical metric and simple.Therefore, $\Delta(X,Y)$ is a similarity measure between two regions X and Y based on the Euclidean distance (4):

$$\Delta(X,Y) = \sqrt{\sum_{e=1}^{512} \left( Hist_X^{\ e} - Hist_Y^{\ e} \right)^2} \qquad (4)$$

Where $Hist_X$ and $Hist_Y$ are the normalized histograms of X and Y. Also, The superscript m is $m_{th}$ member of the histograms. Lower Euclidean distance between X and Y means that the similarity between them is higher.

## 3.2. The Merging Process

In the merging process we have three regions in the image with diferent lables: the object regions denote by $R_O$ that are identified by obtained keypoints from training ASIFT algorithm, the regions around the images denote by $R_B$ that usually are not inclusive objects, therefore, we cover all around the images as initial background regions to help and start the merging process and the third regions, the regions without any sign denote by N.

The merging rule is defined in continue (5). Let Y be an adjacent region of X and $S_Y$ is the set of Y's adjacent region that X is one of them. The similarity between Y and $S_Y$, i.e. $\Delta(Y,S_Y)$, is calculated. We will merge X and Y together if and only if the similarity between them is the maximum among all the similarities $\Delta(Y,S_Y)$.

$$\text{Merge X and Y if } \Delta(X,Y) = \max \Delta(Y,S_Y) \qquad (5)$$

The main strategy is to keep object regions from merging and merge background regions as many as possible. In the first stage, we try to merge background regions with their adjacent regions. Each region $B \in R_B$ will be merged into adjacent region if the merging rule is be satisfied. If the merging accured, the new region has the same label as region B. This stage will be repeated iteratively and in each iteration $R_B$ and N will be updated. Obviously, $R_B$ expands and N shrink due to merging process. The process in this stage stops when background regions $R_B$ can not find any region for new merging.

After this stage, still, some background regions are, that can not be merged due to merging rule. In next stage we will focus on the remaining regions in N from the first stage that are combination of background (N) and object ($R_O$). As before, the regions will be merged based on merging rule and the stage will be repeated iteratively and updated and stops when the entire region N can not find new region for merging. These stages will be repeated again until the merging is not occured. Finaly, all remaining regions in N will be labeled as object and merged into $R_O$ and we can easily extract the object from the image.

## 4. EXPERIMENTAL RESULTS

Our results show that the proposed method is very powerfull and robust for object recognition and detection. We trained three objects (rack, bottle and calendar) from an online dataset [16] in several views with different scales and illuminations for checking our method. Using ASIFT keypoints in the region merging algorithm based on maximal similarity, helped us to obtain significant results. These keypoints are very exact and come from a strong algorithm (ASIFT) with statistical basis. Moreover, the keypoints give us the best information of objects which are very useful for merging process.

Figure 7, shows ASIFT keypoints on the images with initial mean shift segmentation and detected objects with their boundaries. First row, (a), is relevant to the rack dataset, second row, (b), is the bottle and thirth row, (c), is the calendar.

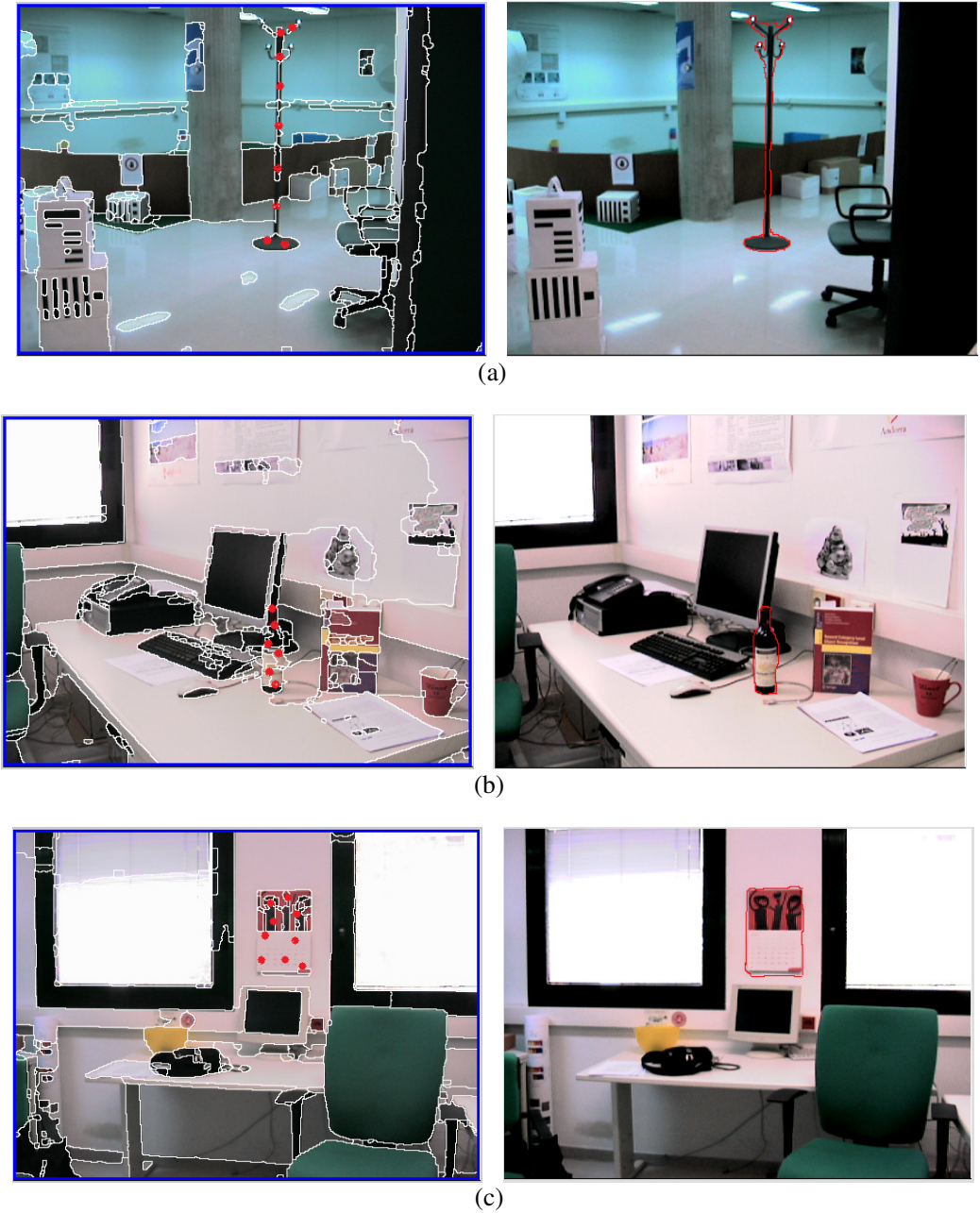

(a)

(b)

(c)

Figure 7. Left column: ASIFT keypoints with initial mean shift segmentation
right column: detected object with its boundary.

Accuracy rate (6) of full boundary detection for each object is computed and shown in Table 1. Results show that the accuracy rate of datasets is close together.

$$Accuracy\ Rate = \ 100\ \times\ \frac{N_{to}}{N_{do}} \qquad (6)$$

Where $N_{to}$ indicates the number of trained images of each dataset and $N_{do}$ means the number of full detected objects respective to the dataset.

Table 1.  Accuracy rate of three datasets.

| Dataset | No. of trained images | No. of tested images | No. of full detected object | Accuracy rate |
|---------|----------------------|---------------------|----------------------------|---------------|
| Rack | 60 | 30 | 26 | 86.66 % |
| Bottle | 60 | 30 | 28 | 93.33 % |
| Calendar | 60 | 30 | 29 | 96.66 % |

In the previous paper [17], we used SIFT keypoints to recognize the objects and utilized  City Block distance as similarity measure to compare the regions. But, in this paper, we have used ASIFT keypoints instead of SIFT keypoints. Also, the Euclidean distance has been replaced with the City Block distance. Comparison between detection rates of [17] and the proposed algorithm in this paper, shows that the proposed method is more accurate than [17] for object recognition and detection. For this Comparison, we have applied the proposed algorithm on 3 used dataset in [17]. The results of the comparison is shown in Table 2.

Table 2.  Comparison between detection rates of [17] and the proposed algorithm.
(The rates are relevant to 3 used dataset in [17])

| Dataset | Detection rate by method in [17] | Detection rate by proposed algorithm |
|---------|----------------------------------|--------------------------------------|
| Book | 91.66 % | **95.83 %** |
| Monitor | 87.5 % | **91.66 %** |
| Stapler | 83.33 % | **91.66 %** |

Also, some of the other experimental results are shown in Figure 8.

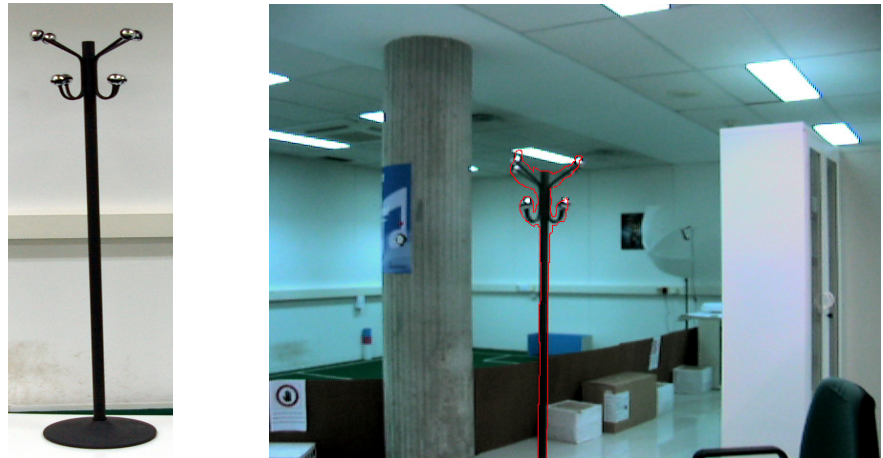

(a)

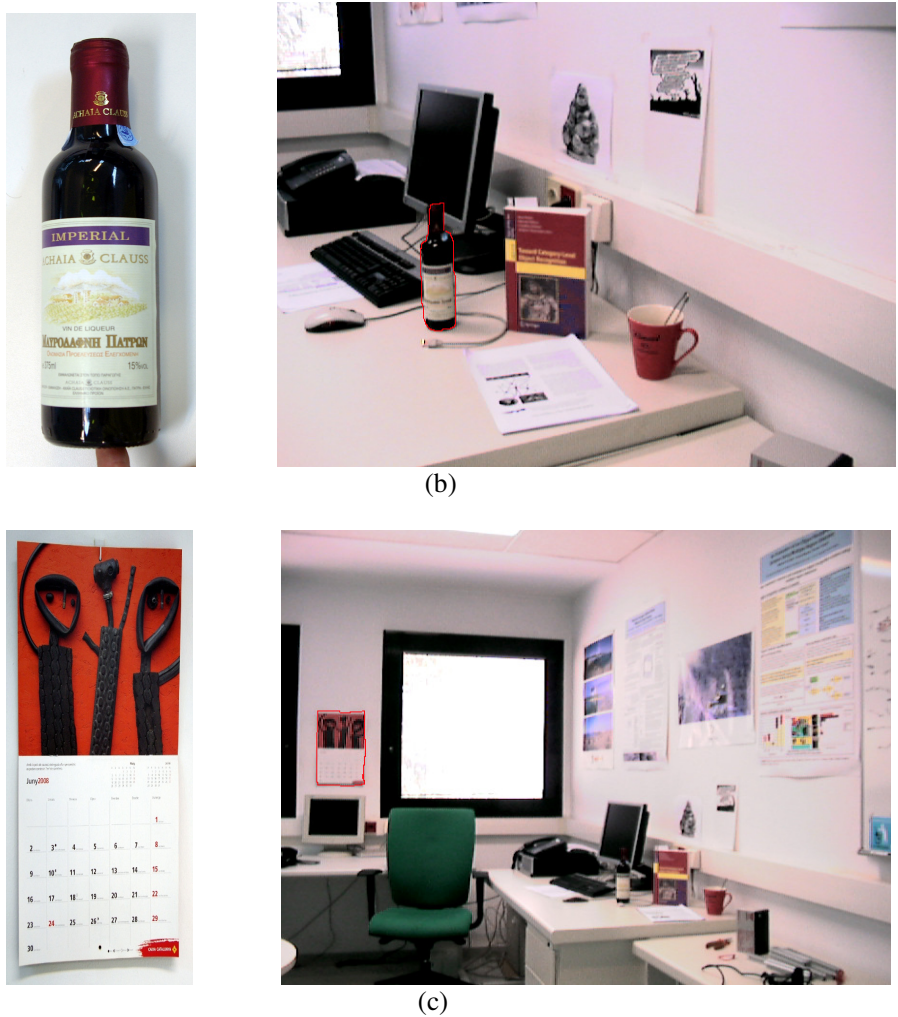

Figure 8. Left column: one of the trained images for an object, right column:
the detected object. (a): rack, (b): bottle (c): calendar.

## 5. CONCLUSIONS

This paper presented an object recognition and detection algorithm. These targets are achieved by combining ASIFT and a region merging segmentation algorithm based on a similarity measure. We train different objects seperately in several images with multiple aspects and camera viewpoints to find the best keypoints for recognizing them in the other images. These keypoints will be applied to the region merging algorithm. The merging process is started by using keypoints and presented similarity measure (Euclidean distance). The regions will be merged based on the merging role and Finaly, the object will be detected well, with its boundary. A final conclusion is that the more keypoints are obtained, and the more accurate they are, the results will be better and more acceptable. Currently, we are working to develop an approach for shape recognition of objects in images. We hope to present a robust algorithm for shape recognition by using the extracted boundaries of objects based on the proposed algorithm in this paper.

# REFERENCES

[1] Jean-Michel Morel, Guoshen Yu, (2009) "ASIFT: A New Framework for Fully Affine Invariant Image Comparison", SIAM Journal on Imaging Sciences, PP: 438-496.

[2] Guoshen Yu, Jean-Michel Morel, (2009) "A Fully Invariant Image Comparison Method", IEEE International Conference on Acoustics, Speech, and Signal Processing (ICASSP), PP: 1597-1600.

[3] David Lowe, (2004) "Distinctive Image Feature Transform from Scale Invariant Keypoints", International Journal of Computer Vision, PP: 91-110.

[4] Shivani Agarawal, Aatif Awan, Dan Roth, (2004) "Learning to Detect Objects in Images via a Sparse, Part-Based Representation", IEEE Transactions on Pattern Analysis and Machine Intelligence, PP: 1475-1490.

[5] Constantine Papageorgiou, Tomaso Poggio, (2000) "A Trainable System for Object Detection". International Journal of Computer Vision, PP: 15-33.

[6] Dingding Liu, Kari Pulli, Linda G. Shapiro, Yingen Xiong, (2010) "Fast Interactive Image Segmentation by Discriminative Clustering", ACM Multimedia Workshop on Mobile Cloud Media Computing, PP: 47-52.

[7] Bo Peng, Lei Zhang, David Zhang, (2001) "Image Segmentation by Iterated Region Merging with Localized Graph Cuts", Pattern Recognition Journal, PP: 2527-2538.

[8] Jifeng Ning, Lei Zhang, David Zhang, Chengke Wu, (2010) "Interactive Image Segmentation by Maximal Similarity based Region Merging", Pattern Recognition Journal, PP: 445-456.

[9] David G. Lowe, (1999) "Object Recognition from Scale Invariant Features", International conference of Computer Vision, PP: 1150-1157.

[10] Mathew Brown, David Lowe, (2002) "Invariant Features from Interest Point Groups", British Machine Vision Conference, PP: 556-565.

[11] Kristian Mikolajczyk, Cordelia Schmid, (2005) "A Performance Evaluation of Local Descriptor", IEEE Transaction on Pattern Analysis and Machine Intelligence, PP: 1615-1630.

[12] Dorin Comaniciu, Peter Meer, (2002) "Mean Shift: A Robust Approach toward Space Analysis", IEEE Transactions on Pattern Analysis and Machine Intelligence, PP: 603-619.

[13] EDISON system. (http://coewww.rutgers.edu/riul/research/code/EDISON/index.html).

[14] Jens Kaftan, Andre Bell, Til Aach, (2008) "Mean Shift Segmentation - Evaluation of Optimization Techniques", International Conference on Computer Vision Theory and Applications, PP: 365-374.

[15] Qiming Luo, Taghi Khoshgoftaar, (2004) "Efficient Image Segmentation by Mean Shift Clustering and MDL-Guided Region Merging", International Conference on Tools with Artificial Intelligence, PP: 337-343.

[16] Online Dataset. (http://www.iiia.csic.es/~aramisa/datasets/iiia30.html).

[17] Reza Oji, Farshad Tajeripour, (2012) "Full Object Boundary Detection by Applying Scale Invariant Features in a Region Merging Segmentation Algorithm", International Journal of Artificial Intelligence and Applications, PP: 41-50.

## Author

**Reza oji** received the B.Sc. degree in Computer Software Engineering from Azad University of Shiraz. He received the M.Sc. degree in Artificial Intelligence from International University of Shiraz. His main research interests include image processing and machine vision.

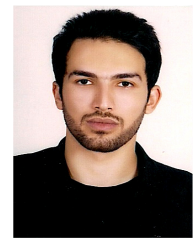